\documentclass[conference]{IEEEtran}
\IEEEoverridecommandlockouts
\usepackage{cite}
\usepackage{amsmath,amssymb,amsfonts}

\usepackage{algorithmic}
\usepackage{graphicx}
\usepackage{textcomp}
\usepackage{xcolor}
\usepackage{soul}
\usepackage{subfigure}
\usepackage{hyperref}
\usepackage[all]{hypcap}
\usepackage{verbatim}
\usepackage{media9}
\usepackage{balance}
\usepackage{makecell}
\usepackage{tabularx}
\usepackage{multirow}
\def\BibTeX{{\rm B\kern-.05em{\sc i\kern-.025em b}\kern-.08em
    T\kern-.1667em\lower.7ex\hbox{E}\kern-.125emX}}

\begin{document}

\bstctlcite{IEEEexample:BSTcontrol}

\title{All On-Board: Fully On-Chip Neuromorphic Q-Learning with Embedded CartPole Simulation\\

\thanks{This work was supported by the US Department of Energy National Nuclear Security Administration’s Office of Defense Nuclear Nonproliferation Research \& Development (DNN R\&D) at Los Alamos National Laboratory under contract 89233218CNA000001.}
}

\author{
    \IEEEauthorblockN{\href{https://orcid.org/0000-0002-2327-7788}{Steven C. Nesbit}\IEEEauthorrefmark{1}\IEEEauthorrefmark{2}, \href{mailto:giovannimichel2024@u.northwestern.edu}{Giovanni T. Michel}\IEEEauthorrefmark{3}\IEEEauthorrefmark{2}, \href{https://orcid.org/0000-0002-4672-9484}{Gerd J. Kunde}\IEEEauthorrefmark{4}, 
    \href{https://orcid.org/0000-0001-5345-3781}{Edward Kim}\IEEEauthorrefmark{1}, and \href{https://orcid.org/0000-0001-8036-6624}{Andrew T. Sornborger}\IEEEauthorrefmark{2}}
    \IEEEauthorblockA{
        \IEEEauthorrefmark{1}College of Computing and Informatics, Drexel University, Philadelphia, PA, USA\\
        \IEEEauthorrefmark{2}Information Sciences (CAI-3), Los Alamos National Laboratory, Los Alamos, NM, USA\\
        \IEEEauthorrefmark{3}Department of Electrical and Computer Engineering, Northwestern University, Evanston, IL, USA\\
        \IEEEauthorrefmark{4}Nuclear \& Particle Physics \& Applications (P-3), Los Alamos National Laboratory, Los Alamos, NM, USA\\
        Email: \IEEEauthorrefmark{2}\href{mailto:nesbitsc@lanl.gov}{nesbitsc@lanl.gov}
    }
}

\maketitle

\noindent
\begin{abstract}
As AI models grow in size and usage, their energy demands increase dramatically, raising sustainability and economic concerns. Neuromorphic hardware, inspired by the energy efficiency of the brain, seeks to address this challenge by offering low-power, fast-processing alternatives to conventional computing. Such hardware is particularly well-suited to control systems deployed in resource-constrained environments, which are best trained via reinforcement learning (RL). This contribution presents the design and implementation of a fully on-chip, closed-loop Loihi~2 RL agent. Our neuromorphic circuit consists of a fully embedded Q-learning algorithm and an on-chip simulation of the CartPole-v0 environment on Loihi~2. Our Q-learning algorithm trained the same number of successful agents as the CPU implementation in only half the execution time and with two orders of magnitude less dynamic power. These findings demonstrate the viability of RL on neuromorphic hardware and highlight its promise for building energy-efficient, real-time, embedded AI systems. 

\end{abstract}


\section{Introduction}

\subsection{Energy Efficiency in Artificial Intelligence}

The growing demand for AI applications has brought about Nobel Prize-winning advancements across various domains such as physics and chemistry \cite{li2024nobel_ai}. However, these advancements come with increasing computational costs, particularly in terms of energy consumption. Energy efficiency in AI models has become an urgent concern, driven by the growing environmental impact of large-scale machine learning (ML) systems and the economic implications of high-power computational infrastructure \cite{strubell2019energy, schwartz2020greenai}. As AI continues to scale, optimizing energy consumption is crucial for both economic sustainability and mitigating the environmental burden \cite{patterson2021carbon, anthony2020carbon, amodei2018ai, bender2021dangers}.

One of the key strategies for addressing the energy challenges in AI is the development of energy-efficient algorithms and architectures. Research in this area has focused on reducing the number of computations required for both training and inference while maintaining or improving model performance \cite{han2015deep}. Several approaches have been proposed to achieve this goal, including model compression, quantization, pruning, and the development of specialized hardware accelerators \cite{hubara2017quantized, jouppi2017tpu}.

The design of energy-efficient hardware accelerators is a critical area of research. GPUs and TPUs have been widely adopted for AI workloads due to their ability to perform parallelized computations \cite{jouppi2017tpu}. However, neuromorphic hardware, which mimics the brain’s event-driven processing capabilities, offers a promising path forward for energy-efficient AI. Neuromorphic systems, such as Intel’s Loihi~2 chip, leverage spiking neural networks (SNNs) to process information in a manner that closely resembles the human brain \cite{davies2018loihi, davies2021advancing, intel2021loihi2}. SNNs operate on discrete events (spikes), which means that computations only occur when a spike is generated, significantly reducing energy consumption \cite{roy2019spikes, davies2021advancing}, while also reducing power density. For instance, the IBM TrueNorth neuromorphic chip power density was $20$ mW/cm$^2$ as opposed to typical CPU power densities of $50 - 100$ W/cm$^2$ \cite{merolla2014million}. Moreover, neuromorphic systems are inherently parallel, allowing them to perform many computations simultaneously with minimal power consumption. This makes them ideal for tasks where low latency and energy efficiency are critical \cite{davies2021advancing, mead1990neuromorphic}.


Despite these advantages, a critical gap remains: implementing complete learning algorithms on neuromorphic hardware without external computational support. Hardware-specific constraints, including limited on-chip plasticity rules, local-only memory access, asynchronous execution, and reduced numerical precision, make on-chip learning algorithms challenging to implement. Exact implementations of conventional learning algorithms remain uncommon; neuromorphic backpropagation has demonstrated that a conventional gradient-based learning algorithm can be executed entirely on Intel's Loihi neuromorphic processor using synfire-gated control of information propagation and learning \cite{spikingbackprop,synfire}.

Neuromorphic reinforcement learning has nevertheless been explored through several complementary approaches. Spiking actor--critic networks have solved simulated control tasks including CartPole \cite{fremaux2013actorcritic}, while memristive systems have demonstrated or modeled in-situ and online reinforcement learning for CartPole and related benchmarks \cite{wang2019memristor,dalgaty2021insitu,vlasov2023memristor}. On Intel Loihi, prior work deployed reinforcement-learning policies or value networks for low-power control, including spiking DDPG for robotic navigation and deep spiking Q-networks for CartPole and Acrobot \cite{tang2020reinforcement,tang2021population,akl2021porting}; related neuromorphic control systems have also demonstrated on-chip adaptation outside the Q-learning setting \cite{vitale2021eventdriven}. More recent work has examined directly trained spiking RL on neuromorphic accelerators, event-based neuromorphic RL architectures, and hardware-feasible spike-based Q-learning \cite{zanatta2023direct,chevtchenko2024neuromorphic,shin2026spike}. These studies establish the feasibility of energy-efficient neuromorphic control and learning, but differ from the present implementation in computational boundary: here, the Q-learning update, action selection, reward/state handling, and the CartPole environment simulation are co-located on Loihi~2 so that the closed training loop does not require host computation between environment transitions.


\begin{figure}
    \centering

    \subfigure[]{%
        \includegraphics[width=0.20\linewidth]
        {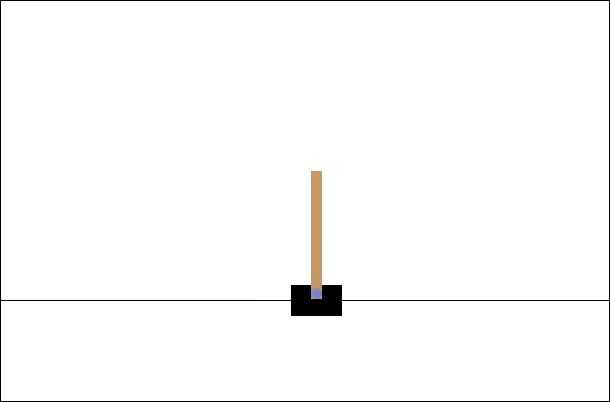}%
        \label{fig:vid_cartpole_python_ideal_q_mat}}
    \hfill
    \subfigure[]{%
        \includegraphics[width=0.20\linewidth]
        {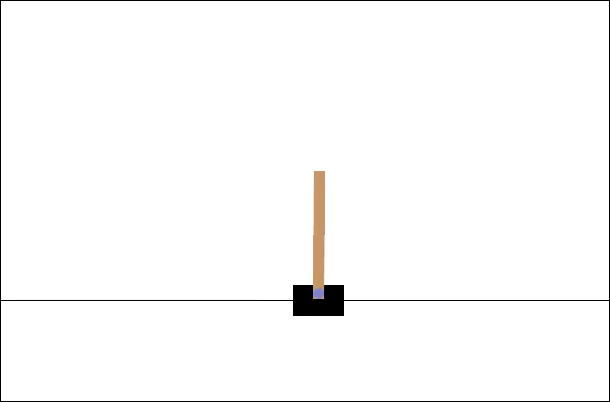}%
        \label{fig:vid_cartpole_loihi_ideal_q_mat}}
    \hfill
    \subfigure[]{%
        \includegraphics[width=0.20\linewidth]
        {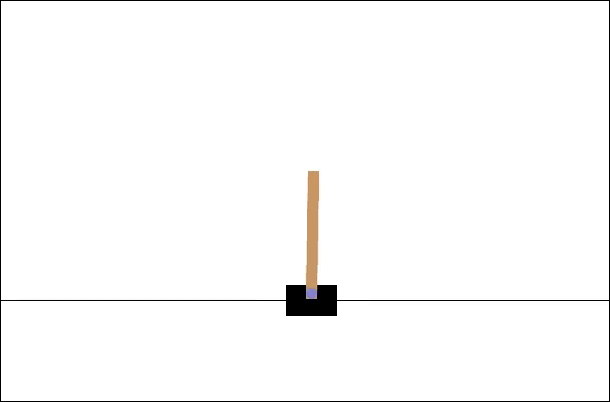}%
        \label{fig:vid_cartpole_python_on_chip_trained_0}}
    \hfill
    \subfigure[]{%
        \includegraphics[width=0.20\linewidth]
        {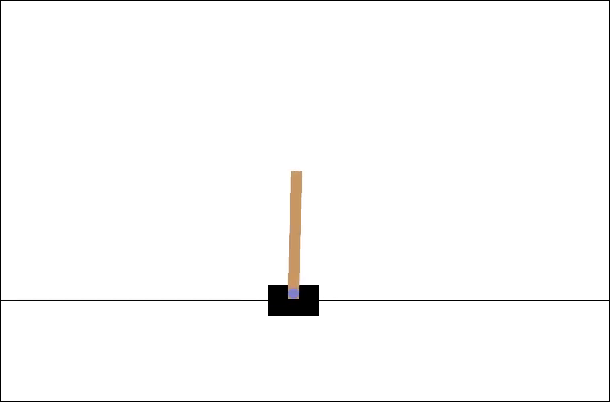}%
        \label{fig:vid_cartpole_loihi_on_chip_trained_0}}

    \medskip

    \subfigure[]{%
        \includegraphics[width=0.20\linewidth]
        {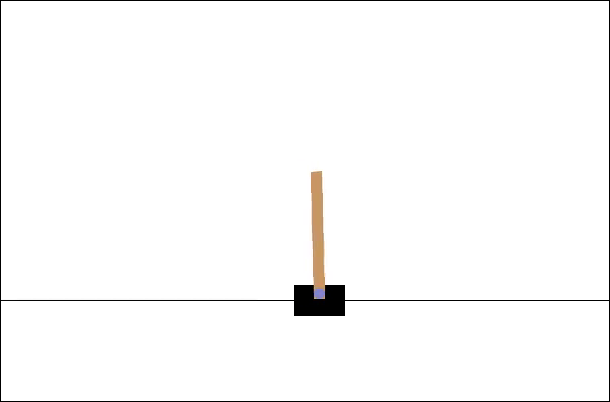}%
        \label{fig:vid_cartpole_python_on_chip_trained_1}}
    \hfill
    \subfigure[]{%
        \includegraphics[width=0.20\linewidth]
        {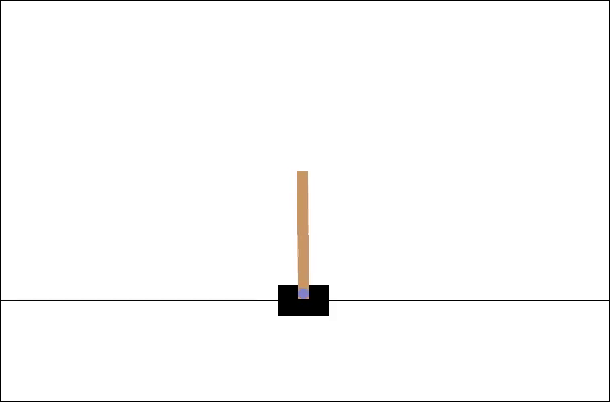}%
        \label{fig:vid_cartpole_loihi_on_chip_trained_1}}
    \hfill
    \subfigure[]{%
        \includegraphics[width=0.20\linewidth]
        {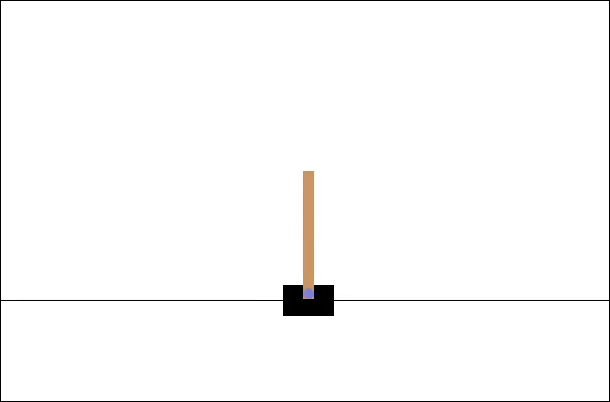}%
        \label{fig:vid_cartpole_python_cpu_trained_0}}
    \hfill
    \subfigure[]{%
        \includegraphics[width=0.20\linewidth]
        {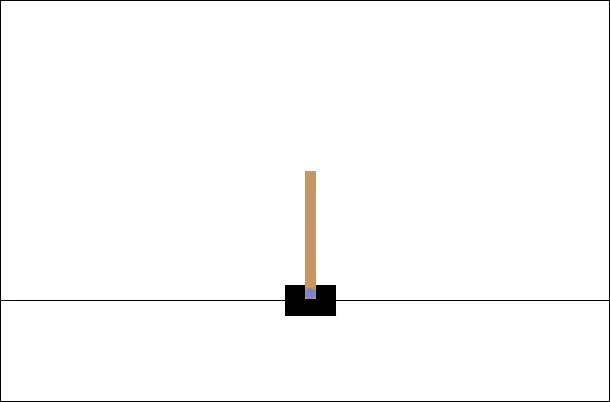}%
        \label{fig:vid_cartpole_python_cpu_trained_1}}

    \medskip

    \subfigure[]{%
        \includegraphics[width=0.20\linewidth]
        {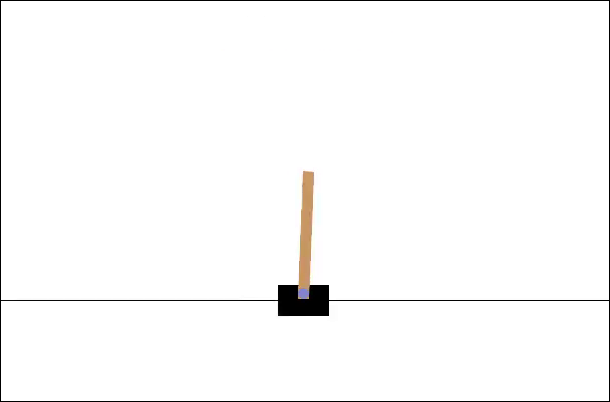}%
        \label{fig:vid_cartpole_loihi_inf_0ts}}
    \hfill
    \subfigure[]{%
        \includegraphics[width=0.20\linewidth]
        {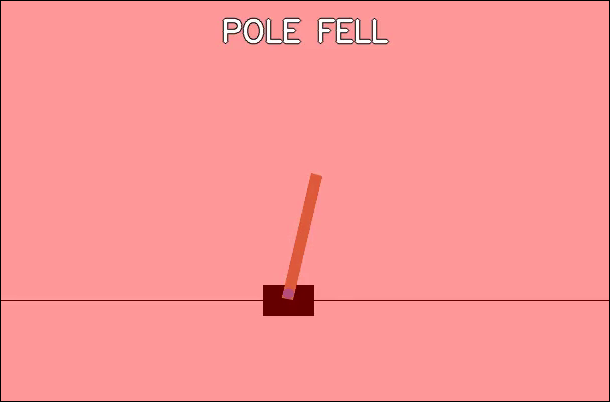}%
        \label{fig:vid_cartpole_loihi_inf_90kts}}
    \hfill
    \subfigure[]{%
        \includegraphics[width=0.20\linewidth]
        {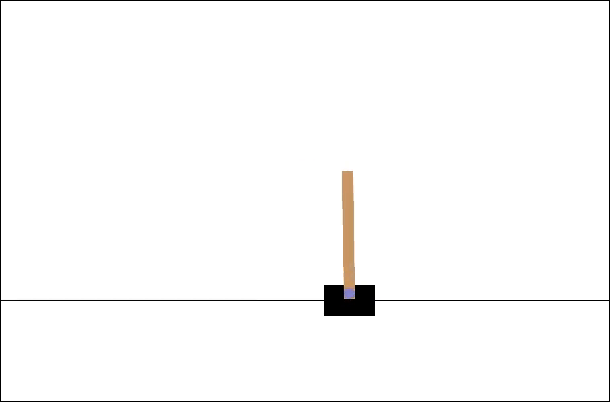}%
        \label{fig:vid_cartpole_loihi_inf_180kts}}
    \hfill
    \subfigure[]{%
        \includegraphics[width=0.20\linewidth]
        {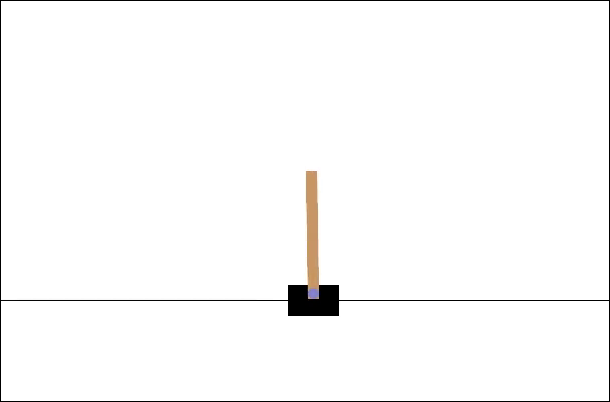}%
        \label{fig:vid_cartpole_loihi_inf_270kts}}

    \medskip

    \subfigure[]{%
        \includegraphics[width=0.20\linewidth]
        {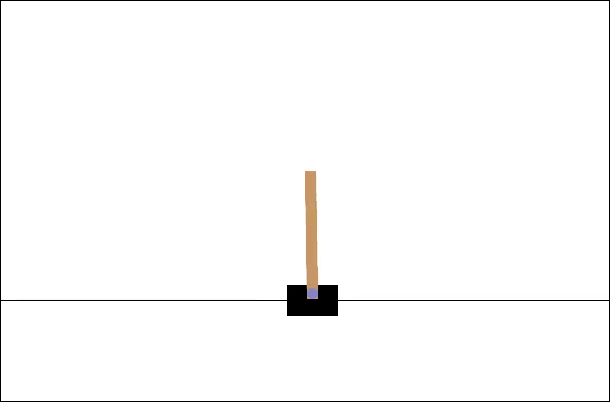}%
        \label{fig:vid_cartpole_loihi_inf_360kts}}
    \hfill
    \subfigure[]{%
        \includegraphics[width=0.20\linewidth]
        {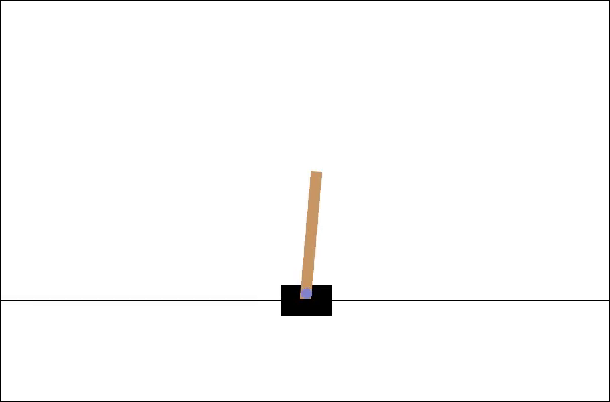}%
        \label{fig:vid_cartpole_loihi_on_chip_0ts}}
    \hfill
    \subfigure[]{%
        \includegraphics[width=0.20\linewidth]
        {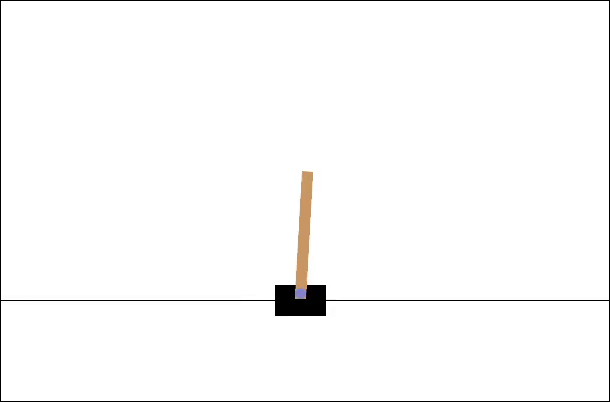}%
        \label{fig:vid_cartpole_loihi_on_chip_90kts}}
    \hfill
    \subfigure[]{%
        \includegraphics[width=0.20\linewidth]
        {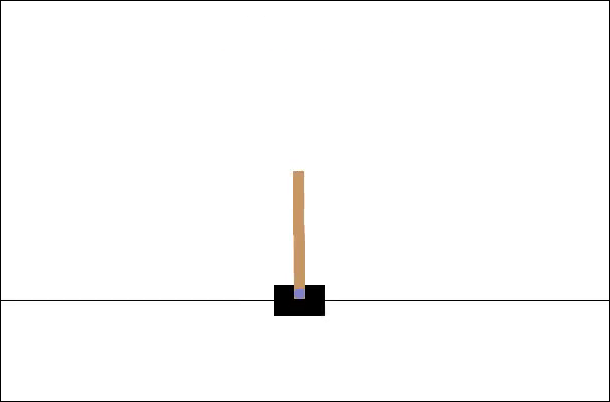}%
        \label{fig:vid_cartpole_loihi_on_chip_180kts}}

    \medskip

    \subfigure[]{%
        \includegraphics[width=0.20\linewidth]
        {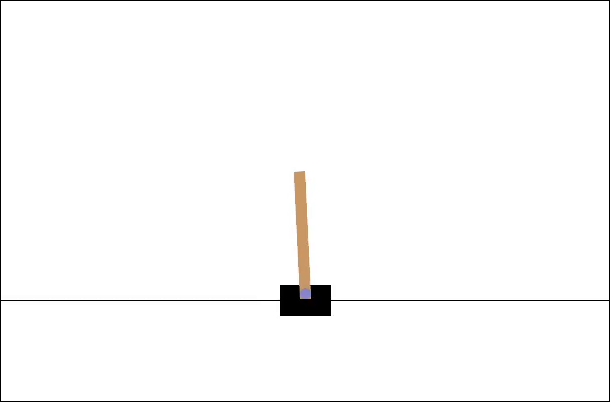}%
        \label{fig:vid_cartpole_loihi_on_chip_270kts}}
    \hfill
    \subfigure[]{%
        \includegraphics[width=0.20\linewidth]
        {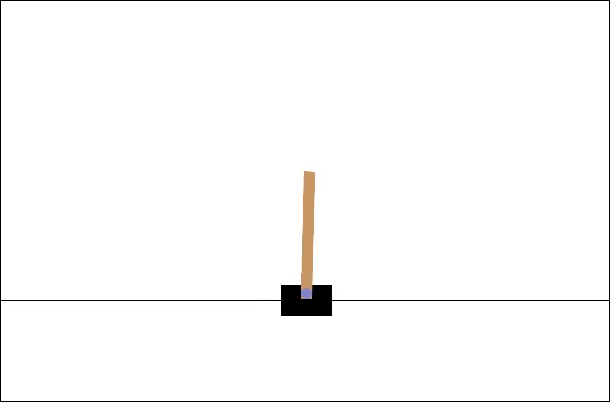}%
        \label{fig:vid_cartpole_loihi_on_chip_360kts}}
    \hfill
    \subfigure[]{%
        \includegraphics[width=0.20\linewidth]
        {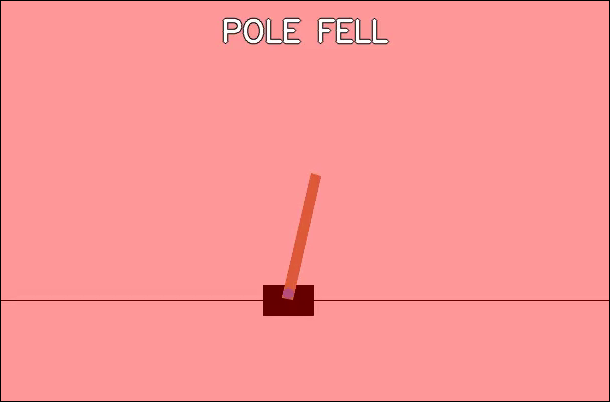}%
        \label{fig:vid_cartpole_loihi_rand_percent_30}}
    \hfill
    \subfigure[]{%
        \includegraphics[width=0.20\linewidth]
        {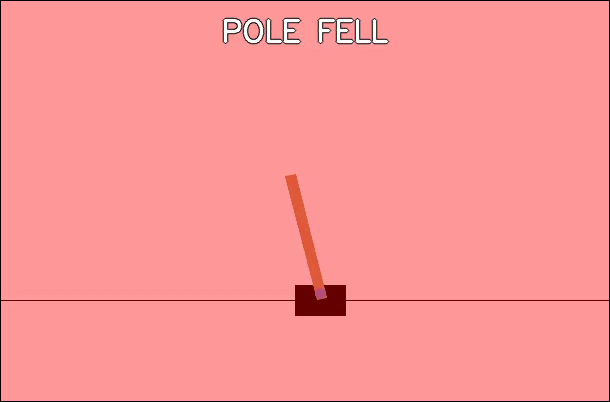}%
        \label{fig:vid_cartpole_loihi_rand_percent_10}}

    \caption{\textbf{Visualizations of CartPole environment.}
    Corresponding 30-second videos are available at
    \href{https://github.com/nesbitsc/all-on-board-figure-1}
    {https://github.com/nesbitsc/all-on-board-figure-1}.
    \textbf{a}:~OpenAI Gym environment inference using agent with ideal Q-table,
    \textbf{b}:~Loihi~2 environment inference using agent with ideal Q-table,
    \textbf{c}:~OpenAI Gym environment inference using on-chip trained agent,
    \textbf{d}:~Loihi~2 environment inference using same agent as (c),
    \textbf{e}:~OpenAI Gym environment inference using on-chip trained agent,
    unique from (c) and (d),
    \textbf{f}:~Loihi~2 environment inference using same agent as (e),
    \textbf{g}:~OpenAI Gym environment inference using CPU-trained agent,
    \textbf{h}:~OpenAI Gym environment inference using CPU-trained agent,
    unique from (g),
    \textbf{i}:~Loihi~2 environment inference using on-chip-trained agent
    trained for 0,
    \textbf{j}:~10\,000,
    \textbf{k}:~20\,000,
    \textbf{l}:~30\,000,
    \textbf{m}:~40\,000 weight updates,
    \textbf{n}:~Loihi~2 environment training at 0,
    \textbf{o}:~10\,000,
    \textbf{p}:~20\,000,
    \textbf{q}:~30\,000,
    \textbf{r}:~40\,000 weight updates,
    \textbf{s}:~Loihi~2 environment inference using exploration rate decay
    function beginning at 30\% and
    \textbf{t}:~10\% randomness.}

    \label{fig:cartpole_videos}
\end{figure}

\subsection{Q-Learning}
\label{sec:q_intro}

One type of algorithm that can be used to train neuromorphic systems for real-time control tasks is Q-learning; a foundational reinforcement learning (RL) algorithm introduced by Watkins in 1989 \cite{watkins1989learning}. Q-learning is a model-free unsupervised learning method that allows an agent to learn the optimal policy for a given environment by interacting with it and maximizing cumulative rewards. Q-learning's compatibility with neuromorphic hardware stems from its discrete state--action space and iterative weight updates, which can be naturally mapped to spike-based neural dynamics. \mbox{Q-learning} has been applied successfully to a wide range of applications, particularly in robotics \cite{kober2013reinforcement, kim2004pathplanning}, autonomous vehicles \cite{shalev2016safe, nguyen2020deep}, and smart grid systems \cite{ernst2006qlearning, bu2019smartgrid}. These applications share a common requirement: real-time learning in resource-constrained environments, where neuromorphic efficiency advantages are most critical.


\begin{equation}
\begin{split}
Q^\mathrm{new}&(s_t, a_t) \leftarrow Q(s_t, a_t) + \\
& \alpha \left[ r_{t+1} + \gamma \max_{a} Q(s_{t+1}, a) - Q(s_t, a_t) \right]
\label{eq:q_learning}
\end{split}
\end{equation}

The Q-learning action--value function \(Q(s_t, a_t)\), which estimates the expected cumulative reward that can be obtained by taking action \(a\) in state \(s\) at time \(t\), is updated via Equation \ref{eq:q_learning}, which may be broken down into the following components:

\begin{equation}
TD_\mathrm{target} = r_{t+1} + \gamma \max_{a} Q(s_{t+1}, a)
\label{eq:target}
\end{equation}

\begin{equation}
TD_\mathrm{error} = TD_\mathrm{target} - Q(s_t, a_t)
\label{eq:error}
\end{equation}

\begin{equation}
U = \alpha \;TD_\mathrm{error}
\label{eq:update}
\end{equation}

\begin{equation}
Q^{\text{new}}(s_t, a_t) \gets Q(s_t, a_t) + U
\label{eq:new_q_value}
\end{equation}

Time step \(t\) refers to the time at which an action was taken rather than the algorithmic time steps described in Section~\ref{sec:q_learning_methods}. The parameter \( \alpha \) is the learning rate, \( r_{t+1} \) is the immediate reward received after taking action \( a_t \) in state \( s_t \), and \( \gamma \) is the discount factor, which determines the importance of future rewards compared to immediate rewards. The term \( \max_{a} Q(s_{t+1}, a) \) is the maximum Q-value for the next state \( s_{t+1} \), representing the best possible outcome from that state onward. \(TD_\mathrm{target}\) is the temporal difference target (an estimate of the future return), \(TD_\mathrm{error}\) is to the temporal difference error (the difference between the agent's current estimate of a state--action value and a more accurate estimate based on observed experience), and \(U\) is the Q-value update term.

\subsection{CartPole Environment}

The CartPole problem is a classical control task widely used as a benchmark for testing the effectiveness of RL algorithms such as Q-learning \cite{barto1983neuron, sutton2018reinforcement}. It was developed by Michie and Chambers in 1968 \cite{michie1968boxes} and has become a fundamental task in the field of machine learning and adaptive control systems. The problem consists of balancing a pole on a moving cart by applying forces to the cart in either direction. The objective is to prevent the pole from falling by keeping it upright for as long as possible, which is achieved by controlling the horizontal movement of the cart. Recent work has formalized CartPole specifically as a neuromorphic-computing benchmark and proposed progressively more difficult variants, noting that the standard formulation provides a useful starting point but is not by itself a particularly challenging test of modern AI agents~\cite{plank2025cartpole}. We intentionally retain the standard CartPole-v0 formulation here because the objective is not to establish state-of-the-art CartPole control performance, but to evaluate whether the complete reinforcement-learning loop, including the learning rule, action selection, reward processing, and nonlinear environment dynamics, can execute on a single neuromorphic processor.

The environment's dynamics, governed by classical mechanics equations involving trigonometric functions and divisions, present a cart that moves along a horizontal track, with a pole that pivots freely along the cart's center. The state of the system can be described by two variables: the position, \( x \), of the cart along the track and the angle, \( \theta \), of the pole relative to the vertical axis. The control input to the system is the horizontal force applied to the cart. This force causes the cart to accelerate, which in turn affects the motion of the pole. The goal is to control the system such that the pole remains balanced while the cart remains within predefined limits on the track. The challenge arises from the fact that small deviations in \( \theta \) can quickly lead to instability, requiring rapid corrective actions to prevent the pole from falling.

\begin{figure}[tbh]
    \centering
    \includegraphics[width=0.8\linewidth]{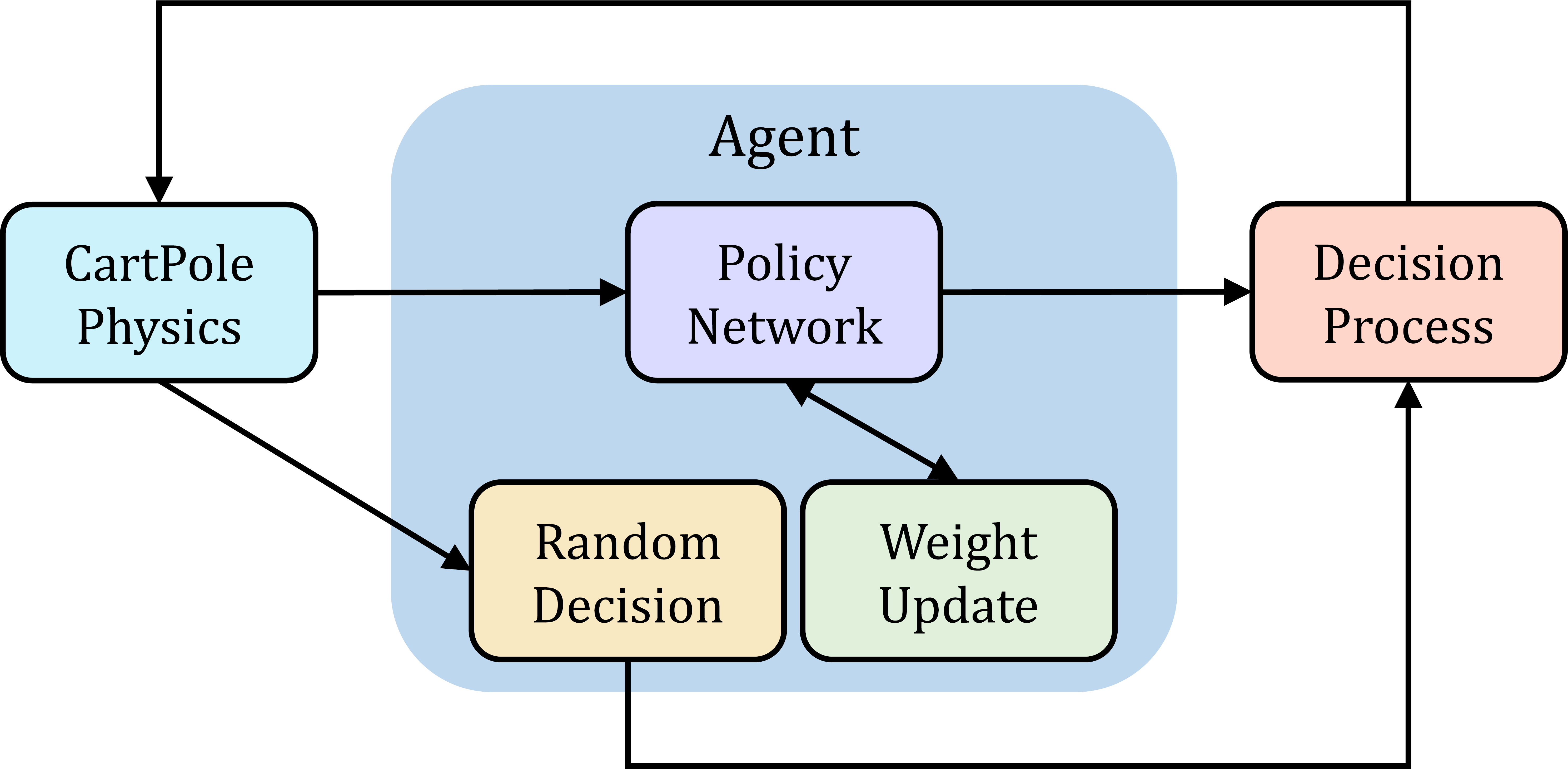}
    \caption{\textbf{Overview of neuromorphic Q-learning circuit architecture.} Angle and angular velocity measurements are sent from the CartPole neuron to the policy network, which sends these values through the Q-table. Q-table output from the policy network goes to the decision process module, which determines whether or not to take a left or right action. The CartPole neuron also sends reward information to the random decision module, which decides whether (exploration) or not (exploitation) to select an action randomly, and if so, makes a random decision and overrides the greedy decision made by the decision process module. The policy network also sends its Q-table output values to the weight update module, which in turn sends Q-value update information to the policy network. Finally, the action selected by the decision process module is sent to the CartPole neuron to calculate the next state of the CartPole environment.}
    \label{fig:overview}
\end{figure}

\section{Methods}

\subsection{Neuromorphic CartPole Simulation}

To enable fully on-chip execution of Q-learning and inference for the CartPole environment, we developed a neuromorphic simulation of OpenAI Gym’s CartPole-v0 for deployment on the Loihi~2 neuromorphic processor utilizing Lava on Loihi microcode. This simulation was adapted from the original OpenAI Gym CartPole source code available on GitHub~\cite{openai_gym_cartpole}. 

\begin{equation}
\ddot{\theta} = \frac{g \sin\theta + \cos\theta \left( -F - m_p l \dot{\theta}^2 \sin\theta \right) / (m_c + m_p)}{l \left( \frac{4}{3} - \frac{m_p \cos^2\theta}{m_c + m_p} \right)}
\label{eq:theta_acceleration}
\end{equation}

\begin{equation}
\ddot{x} = \frac{F + m_p l \left( \dot{\theta}^2 \sin\theta - \ddot{\theta} \cos\theta \right)}{m_c + m_p}
\label{eq:cart_acceleration}
\end{equation}

Equations~\ref{eq:theta_acceleration} and~\ref{eq:cart_acceleration} governing the dynamics of the CartPole system, as outlined by \cite{florian2007correct}, describe the relationship between the applied force, \( F \) ($\pm10$ N), the acceleration, \( \ddot{x} \), of the cart of mass \( m_c \) (1 kg), and the angular acceleration, \( \ddot{\theta} \), of the pole of mass \( m_p \) (0.1 kg) and length \( 2l \) (1 m). CartPole serves as an ideal testbed because: (1) it is a standard RL benchmark enabling direct comparison, (2) its physics simulation can be embedded on-chip, and (3) its discrete action space aligns well with neuromorphic architectures.

\subsubsection{Function Approximation}

The calculations for CartPole dynamics, expressed in Equations~\ref{eq:theta_acceleration} and~\ref{eq:cart_acceleration}, had to be implemented using only a limited microcode instruction set. Notably, the sine and cosine functions were not directly supported and therefore required a custom implementation. The sine function was approximated within the range of $[-0.418,0.418]$~rad (the maximum possible range of the pole angle) by $y = 0.98214x$ with a maximum absolute error of 0.00229. The cosine function was approximated within this same range with the function $y = -0.00012052x^2 + 4096$ with a maximum absolute error of 0.00020. In addition, Newton’s method was employed to iteratively approximate division by finding the roots of the reciprocal function, $\frac{1}{b}$, where $b$ is the divisor, and then multiplying the result by the numerator, giving a maximum absolute error of 0.00291. These calculation errors combine to give a maximum error in $\ddot{\theta}$ of 0.03632 rad/s\textsuperscript{2} and a maximum error in $\ddot{x}$ of 0.00166 m/s\textsuperscript{2}.

\subsubsection{Random Number Generation}

In the OpenAI Gym CartPole environment, random numbers are used to initialize the state variables---cart position, cart velocity, pole angle, and pole angular velocity---uniformly within the range $[-0.05, 0.05]$. These values are generated using the \texttt{numpy.random.uniform()} function, which relies on the Mersenne Twister MT19937 algorithm~\cite{mersenne_twister} and is seeded with nondeterministic data from the operating system~\cite{numpy_generator,numpy_random}. Although replicating the MT19937 algorithm would have provided an exact match to the Python CartPole implementation, it was infeasible to incorporate within the CartPole process on Loihi~2 due to hardware constraints. To address this, a simpler linear congruential generator (LCG)~\cite{lehmer1951} was chosen and is defined by:

\begin{equation}
X_{n+1} = (aX_n + c) \bmod m
\label{eq:lcg}
\end{equation}

\noindent where $X_n$ is the current seed, $a$ is the multiplier, $c$ is the increment, and $m$ is the modulus. To generate pseudo-random 23-bit unsigned integers, the constants $a = 65\,793$, $c = 4\,282\,663$, and $m = 2^{23}$ were used, following the LCG implementation in the \texttt{cc65} development package~\cite{cc65_rand,rng_comparison}. While the LCG produces lower-quality random numbers (period of 2\textsuperscript{23} vs. 2\textsuperscript{19937}), the random number generator was only called a maximum of 400\,000 times by the CartPole process per training run. 

\subsubsection{CartPole State Variables}

In addition to computing cart acceleration and pole angular acceleration, the neuromorphic CartPole simulation maintained and updated four key state variables governing the environment’s physics: cart position, cart velocity, pole angle, and pole angular velocity. These variables were updated at each simulation step using the Euler method, with a fixed time step duration of 0.02~s. After each update, the simulation checked if the environment had entered a failure state. A failure state was triggered if the cart position exceeded $\pm2.4$~m from the center or if the pole angle magnitude surpassed 0.2095~rad. At each time step, the CartPole process output five variables as 24-bit signed integers: pole angle, pole angular velocity, reward, cart position, and cart velocity. Comparisons between neuromorphic and CPU CartPole environments can be seen in Figs.~\ref{fig:vid_cartpole_python_ideal_q_mat} through \ref{fig:vid_cartpole_loihi_on_chip_trained_1}. Further visualizations, described in the caption and below may be seen in Figs~\ref{fig:vid_cartpole_python_cpu_trained_0} through \ref{fig:vid_cartpole_loihi_rand_percent_10}.

\begin{figure*}[tbh]
    \centering
    \includegraphics[width=0.95\textwidth]{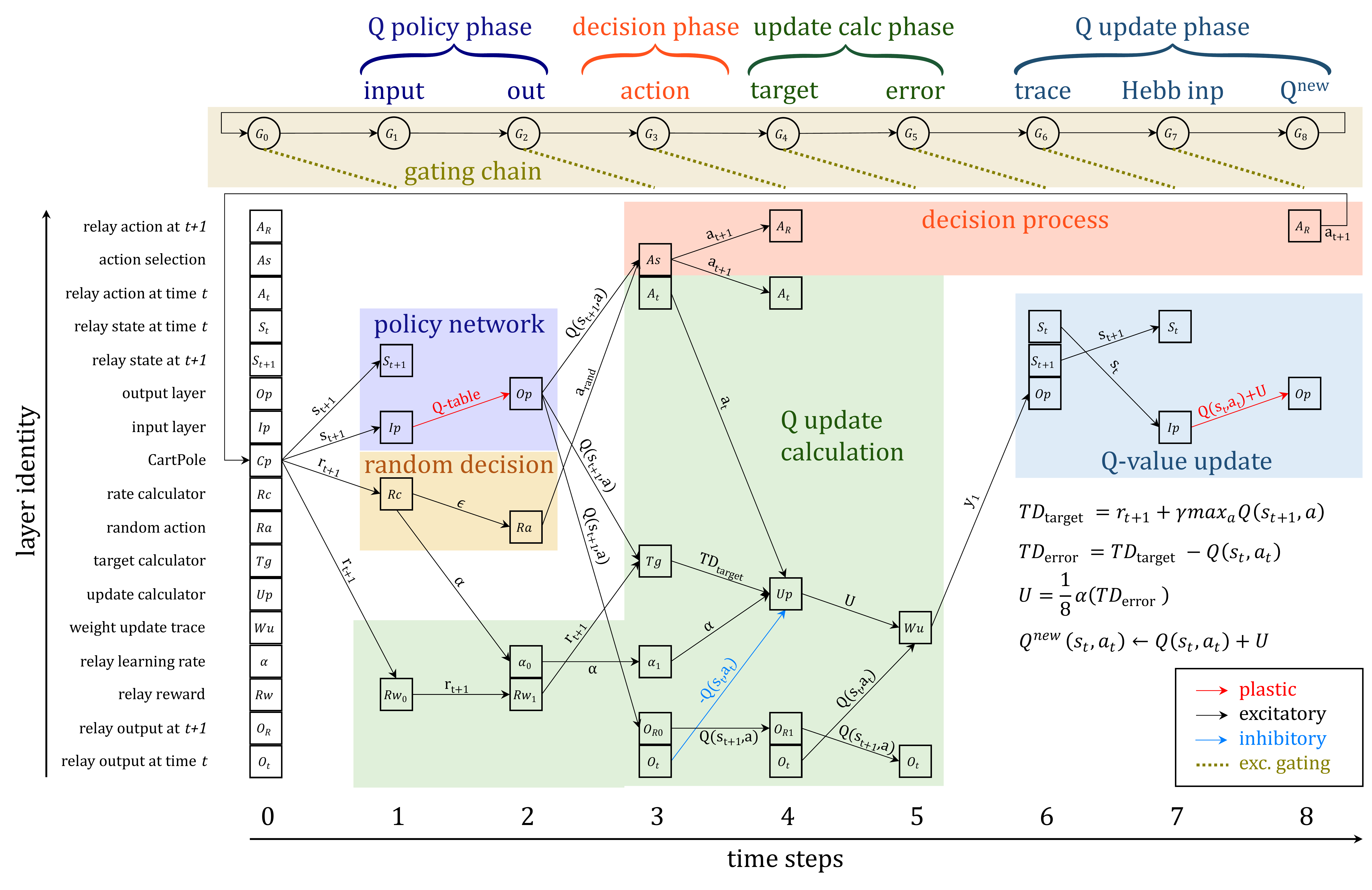}
    \caption{\textbf{Functional connectivity of the neuromorphic Q-learning circuit.} Layers are only shown when gated `on,' and synapses are only shown when their target is gated on. The names of the neuron layers are given on the left margin so that the row corresponds to layer identity. The columns correspond to the time steps of the algorithm, which correlate to the time steps on Loihi~2. Arrow labels indicate information flow through the synapse. Variables $\epsilon$ and $\alpha$ refer to the exploration and learning rates, respectively (see Equation~\ref{eq:epsilon_decay}). $a_{\rm{rand}}$, random action; $y_1$,~constrained update term (see Sections~\ref{sec:policy} and \ref{sec:weight_update}).}
    \label{fig:q_learning_diagram}
    \vspace{0cm}
\end{figure*}

\subsection{Q-Learning}
\label{sec:q_learning_methods}

An overview of the neuromorphic Q-learning circuit is shown in Fig.~\ref{fig:overview}. A detailed schematic of functional connectivity is provided in Fig.~\ref{fig:q_learning_diagram}, with hyperparameters outlined in Table~\ref{tab:hyperparameters}. Our neuromorphic Q-learning circuit consists of five interconnected modules, synchronized by a gating chain that controls the nine-step algorithmic loop summarized as follows:

\begin{itemize}

\item \textbf{T\textsubscript{0}}: The CartPole simulator is denoted as layer $Cp$ and updates its state (position, velocity, pole angle, and angular velocity) based on the previous iteration (at the beginning of an epoch, it is set to a random state). $Cp$ gates its angle and angular velocity values to layer $S_{t+1}$ (this value persists in the layer and is used at \textbf{T\textsubscript{6}}) and to layer $Ip$ (the input to the Q-table). $Cp$ gates the reward value (\(+1\) for a balanced pole and in-bounds cart, or \(-1\) otherwise) to rate calculator layer $Rc$ and a relay layer, $Rw_0$. The first neuron in the gating chain triggers processing in \textbf{T\textsubscript{1}} (and similarly downstream).

\item \textbf{T\textsubscript{1}}: The input layer encodes the state as a one-hot vector of length 72 and sends it through a weight matrix (\mbox{Q-table}) to the output layer, $Op$. Simultaneously, the rate calculation process $Rc$ uses the reward to compute the episode score and derives the exploration rate \(\epsilon\) and learning rate \(\alpha\) using Equation~\ref{eq:epsilon_decay}. These values are then sent to the random action process, $Ra$, and a relay neuron, $\alpha_0$, respectively. The episode score is also used to determine whether or not to terminate training early. If an episode score of 500 is reached, $Rc$ will emit an inhibitory signal to the gating chain at neuron \(G_2\) to halt the execution of the synfire gating chain, and thus end training. Additionally, the relay value from $Rw_0$ is gated to $Rw_1$ at this step.

\item \textbf{T\textsubscript{2}}: The output layer, $Op$, (of size two, corresponding to left and right actions) then sends the Q-values for state--action pair \((s_{t+1}, a)\) to the action selection layer, $A_s$, the target layer, $Tg$, and a relay neuron, $O_{R0}$, that stores \(Q(s_t, a_t)\) for the next loop’s weight update. $Rw_1$ gates the reward (scaled by a weight of 8, yielding values of \(-8\) or \(+8\)) to target layer $Tg$. The random action process can also send an override signal to the action selection layer, $As$, to initiate a random action. Relay neuron $\alpha_0$ gates the learning rate $\alpha$ to layer $\alpha_1$.

\item \textbf{T\textsubscript{3}}: The selected action \(a_{t+1}\) stored in $As$ is gated to a relay layer, $A_R$, for later use by the CartPole layer, $Cp$, and is gated to $A_t$ for use in the next iteration. $A_t$, persistent from the previous iteration, is gated to the update calculator layer, $Up$. Layer $Tg$ computes $TD_\mathrm{target}$ from its input via Equation~\ref{eq:target} and gates it to $Up$. Layer $\alpha_1$ gates $\alpha$ to layer $Up$, and the Q-value at $t$ is gated from $O_t$ to layer $Up$. $O_t$ contains persistent policy output from the previous iteration. Simultaneously, $O_{R0}$ is gated to $O_{R1}$.

\item \textbf{T\textsubscript{4}}: The update calculator, $Up$, computes \( \alpha(TD_\mathrm{error})\) (see Equations~\ref{eq:error} and \ref{eq:update}) from its inputs and gates it to weight update trace layer, $Wu$. In parallel, $O_t$ is gated to $Wu$.

\item \textbf{T\textsubscript{5}}: The weight update trace layer, $Wu$, determines the Q-value update, $y_1$, which is gated to $Op$.

\item \textbf{T\textsubscript{6}}: The persistent state in relay neuron $S_t$ is updated and the previous state is sent to the input layer, $Ip$. 

\item \textbf{T\textsubscript{7}}: The input layer, $Ip$, sends \(s_t\) to the output layer, $Op$.

\item \textbf{T\textsubscript{8}}: Via Hebbian learning, the Q-values associated with the current state are updated using the update trace, \(y_1\). The stored action \(a_{t+1}\) is gated as input to the CartPole simulation. Meanwhile, synfire gating chain neuron \(G_8\) sends a signal to \(G_0\), restarting the sequence for the next state--action cycle.

\end{itemize}

\begin{table}[htbp]
\caption{Hyperparameters used in neuromorphic Q-learning.}
\begin{center}
\begin{tabular}{|l|l|}
\hline
\textbf{Hyperparameter} & \textbf{Value} \\
\hline
Q-value update function & Equation~\ref{eq:q_learning} \\
$\alpha$, $\epsilon$ decay function & Equation~\ref{eq:epsilon_decay} \\
reward $r_{t+1}$ & $\pm 8$ \\
discount factor $\gamma$ & $0.99$ \\
Q-table shape & $2 \times 72$ \\
Q-value initialization & $-127$ \\
Q-value range & $[-127,126]$ \\
termination condition & score $= 500$ \\
\hline
\end{tabular}
\label{tab:hyperparameters}
\end{center}
\end{table}

The modules and custom neuron processes shown in Figs. \ref{fig:overview} and \ref{fig:q_learning_diagram} are described in detail for the remainder of this section.

\subsubsection{Gating Chain}

Depicted at the top of Fig.~\ref{fig:q_learning_diagram}, the gating chain serves a similar role to that described in neuromorphic backpropagation~\cite{spikingbackprop}. The gating chain fires in a continuous loop, with each neuron activating the next in sequence, controlling the propagation of information throughout the network. A total of nine gating neurons are used, corresponding to the nine-time step loop of the Q-learning algorithm.

\subsubsection{Policy Network}
\label{sec:policy}

The policy network consists of three unique processes, the first of which is the input layer, $Ip$. The input layer is responsible for receiving pole angle and pole angular velocity information from the CartPole process, $Cp$, and converts this into a one-hot-encoded state vector. This is achieved by discretizing the pole angle into six bins and the pole angular velocity into twelve bins. This discretization was based on empirical testing of quantized off-chip Q-learning. The input layer contains 72 neurons, each corresponding to a unique combination of threshold values for pole angle and pole angular velocity. This 72-state representation balances state space coverage with neuromorphic memory limitations.

The output layer, $Op$, consists of two neurons that are connected to the input layer via a plastic connection with a weight matrix $Q \in \mathbb{R}^{2 \times 72}$, which serves as the Q-table. With the input vector having a value of 1 for the neuron representing the current state, the resulting output vector $q = s Q^\top$ produces both Q-values corresponding to that state. The output layer also receives input indicating how to update the plastic connection. We implement Q-table updates using Hebbian learning, where the \texttt{tag} variable $T$ tracks the target weight change, and the weight update $dW$ applies this change only to the active state--action pair. Specifically:

\begin{equation}
dT = 2u_0y_1 - u_0T
\label{eq:dt}
\end{equation}

\begin{equation}
dW = x_0T
\label{eq:dw}
\end{equation}

In Equation~\ref{eq:dt}, $T$ represents a register in the Loihi~2 learning engine referred to as the \texttt{tag} variable. This 8-bit signed register is updated before the 8-bit signed weight variable $W$, which is updated according to Equation~\ref{eq:dw}. The variable $u$ in Equation~\ref{eq:dt} denotes how frequently to read from specific registers, which is every time step in this case. The variable $y_1$ in Equation~\ref{eq:dt} dictates the value of the weight change. The term $-u_0T$ ensures that the \texttt{tag} variable is always equal to the value of the weight update trace $y_1$. In Equation~\ref{eq:dw}, the variable $x_0$ represents the spike count from input neurons that had sent activity through the plastic connection. This register is multiplied by the \texttt{tag} variable, and the result is bit-shifted right by 1 bit to yield the final weight update $dW$.

The third type of microcoded neuron process used in the policy network is the relay neuron, $S_{t+1}$, which serves as a neuromorphic memory register with data accessible to other neurons. Here it is used to store state information at time $t+1$ for use in Hebbian weight updates, as shown in Fig.~\ref{fig:q_learning_diagram}. 

\subsubsection{Random Decision}

The Q-learning algorithm relies on random action selection to explore the state space and determine which actions yield higher cumulative rewards for given states. The probability of overriding the Q-policy decision with a random choice is governed by Equation~\ref{eq:epsilon_decay}, as used by \cite{muetsch2020cartpole}. The output of this equation is also used as the learning rate in the weight update module. The random decision module in the neuromorphic Q-learning algorithm is responsible for determining when to take a random action, selecting which random action to take, and calculating the learning rate to be used by the weight update module.

\begin{equation}
\epsilon,\alpha = \max \left( 0.1, \min \left( 1, 1.0 - \log_{10} \left( \frac{score + 1}{25} \right) \right) \right)
\label{eq:epsilon_decay}
\end{equation}

The rate calculator process, $Rc$, receives input from the CartPole simulation’s reward value and accumulates the number of valid states as a total score for each episode in the CartPole environment, which serves as input to Equation~\ref{eq:epsilon_decay}. The resulting exploration rate $\epsilon$ is sent from the rate calculator neuron to the random action neuron, $Ra$, so that it can determine the probability of selecting a random action and, subsequently, which action to take. The random action neuron employs the same LCG as the CartPole process. If the random number is greater than the $\epsilon$ input, no random action is selected, otherwise a random value of $-1$ or $1$ is sent as output.

\subsubsection{Decision Process}

The decision process module consists of one unique process, the action selection neuron, $As$, as well as a gated relay neuron, $A_R$, used to store the value sent by the action selection neuron. The action selection neuron receives input from the output layer of the policy network. The weight of the connection from the first neuron of the output layer to the action selection process is $-1$, and the weight of the connection from the second neuron is $1$. These values are summed within the action selection neuron's dendrite accumulator. Therefore, a larger Q-value for the left action results in a negative spiking input to the action selection neuron, a larger Q-value for the right action produces a positive input, and equal Q-values result in an input of 0. In the case of equal Q-values, the action selection neuron randomly selects a left or right action. The action selection neuron also receives random decision input from the random action neuron. Receiving this input overrides both the spiking input from the output layer and the internally generated random decision of the action selection neuron.

\subsubsection{Weight Update}
\label{sec:weight_update}

The weight update module determines how much to adjust the Q-table values and sends this weight update trace, $y_1$, to the output layer. The weight update equation used in this training algorithm is the same as the Q-learning update rule in Equation~\ref{eq:q_learning}---with the learning rate scaled by a factor of $\frac{1}{8}$ to prevent the \( \max_{a} Q(s_{t+1}, a) \) term from overflowing within the allotted 8 bits of precision.

\(TD_\mathrm{target}\) is calculated by the target calculator process, $Tg$, using Equation~\ref{eq:target}. In order to determine the maximum Q-value of the current state--action pair, $\max_{a} Q(s_{t+1}, a)$, the neuron receives distinct inputs corresponding to the Q-values of the left and right actions. The maximum of these two inputs is computed using a simple function available in Lava microcode.

The target calculator neuron also receives the reward input $r_{t+1}$ from a relay neuron, $Rw_1$, downstream of the CartPole neuron's reward output. For a positive reward value, $r_{t+1}$ is added to $\gamma \max_{a} Q(s_{t+1}, a)$, whereas for a negative reward value, only $r_{t+1}$ is sent as output. This is because the expected maximum future reward following a failure state, $\max_{a} Q(s_{t+1}, a)$, is assumed to be 0; thus, \(TD_\mathrm{target}\) simplified to $TD_\mathrm{target} = r_{t+1} + 0$ in such cases. The reward input also serves as a gating signal for the neuron's activity. If no reward input is received, then the neuron produces no output. Otherwise, $TD_\mathrm{target}$ is sent to the update calculator layer, $Up$.

The update calculator process is responsible for performing the calculations shown in Equations~\ref{eq:error} and \ref{eq:update}. To do this, it receives three distinct types of input. The first input to the neuron consists of the summation of the $TD_\mathrm{target}$ value sent from the target calculator neuron and $-Q(s_t, a_t)$, which is transmitted by a gated relay neuron, $O_t$, used to store the state--action Q-values from time~$t$. As a result, the quantity received at this input is $TD_\mathrm{target} - Q(s_t, a_t)$. The neuron also receives learning rate, $\alpha$, input from a relay neuron, $\alpha_1$. This input gates the activity of the update calculator neuron, as the learning rate is multiplied by the $TD_\mathrm{target} - Q(s_t, a_t)$ term and $\frac{1}{8}$ to compute $U$. If no learning rate input is provided, the resulting value of $U$ would be 0. The third type of input received by the update calculator process is the action selection at time $t$ from relay neuron $A_t$. The update calculator process consists of two neurons, which allow it to send different update values for each Q-value associated with a given state.

The final microcoded process to be discussed is the weight update trace process, $Wu$, which consists of two neurons. The first input received by both neurons in the weight update trace process is the output from the left update calculator neuron added to the existing Q-value for the leftward action associated with state~$s_t$. This quantity can be represented as $U_0 + Q(s_t, a_0)$. The second input received by both neurons is the output from the right update calculator neuron added to the existing Q-value for the rightward action, or $U_1 + Q(s_t, a_1)$. Relay neuron $O_t$, downstream of the output layer, is used to store the Q-values output at time~$t$. This neuron is connected to the weight update trace process via a one-to-one connection. Thus, for the third input, each weight update trace neuron individually receives a Q-value corresponding to state~$s_t$ and either action $a_0$ or $a_1$. The updated Q-value is constrained to the 8-bit signed range using the equation:

\begin{equation}
Q^{\text{new}}(s_t, a_t) \gets \min\left(\max\left(Q^{\text{new}}(s_t, a_t), -127\right), 126\right)
\label{eq:constraint}
\end{equation}

This range was chosen to be slightly narrower than the full 8-bit signed integer range $[-128,127]$ to prevent potential overflow resulting from stochastic rounding. Once the updated Q-value is bounded, the weight update trace neuron checks if the resulting left and right Q-values are equal. If so, and if both saturate at $126$, then the updated Q-value is set to $126$ and the opposing Q-value is adjusted to $125$ to maintain a difference. The trace values $y_1$ are then computed as the difference between the new and previous Q-values. These are sent to the output layer to drive Hebbian learning according to the update Equations~\ref{eq:dt} and~\ref{eq:dw}.

\begin{figure}
  \begin{center}
    \textbf{\footnotesize Training Success Rate}\\[0.01cm]
    \makebox[\linewidth][l]{\hspace{2.2cm} \textbf{\footnotesize CPU} \hspace{2.95cm} \textbf{\footnotesize Loihi~2}}\\[-.15cm]
    \subfigure[]{%
      \includegraphics[width=0.43\linewidth]{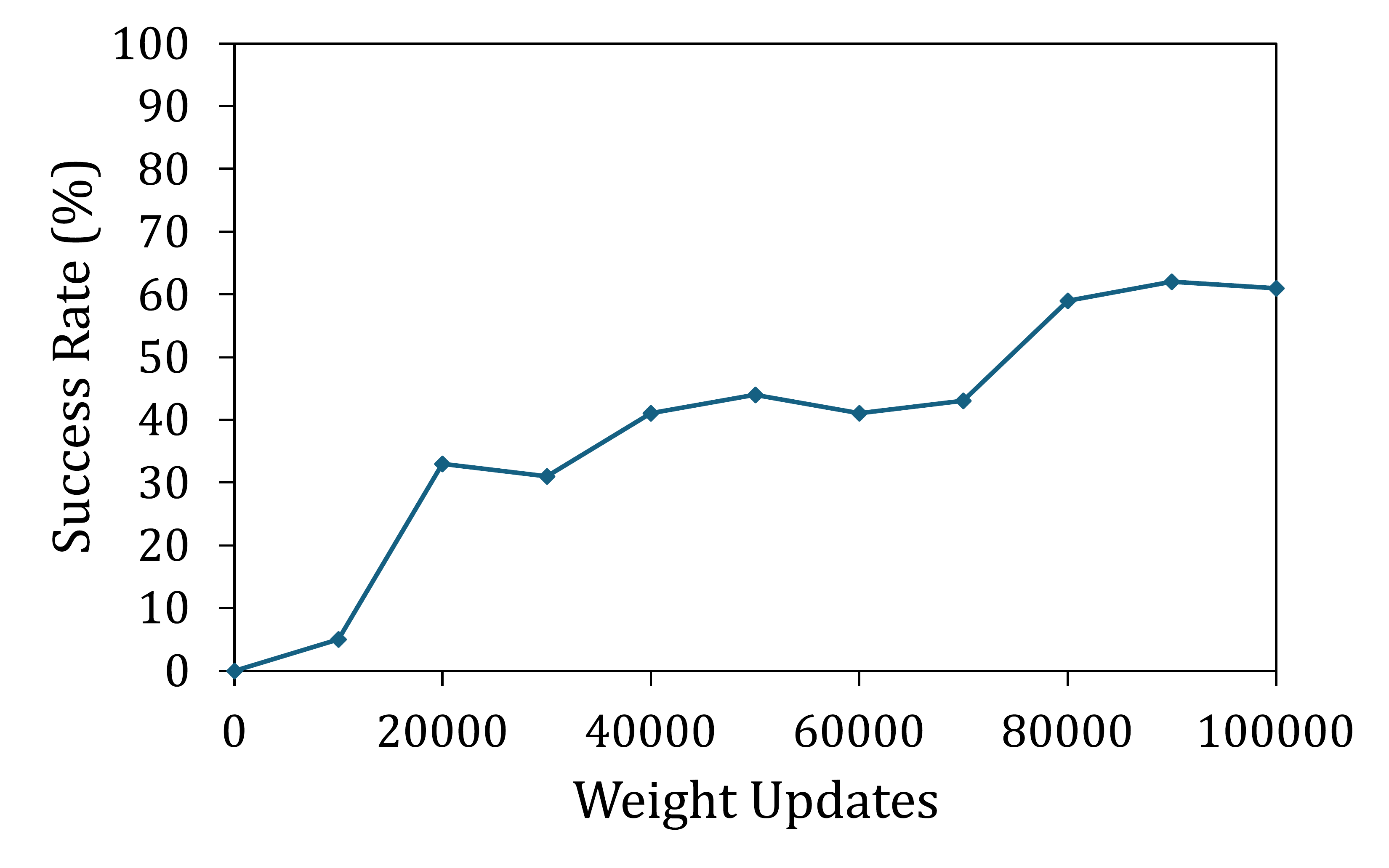}%
      \label{fig:cpu_success_rate}}
    \subfigure[]{%
      \includegraphics[width=0.43\linewidth]{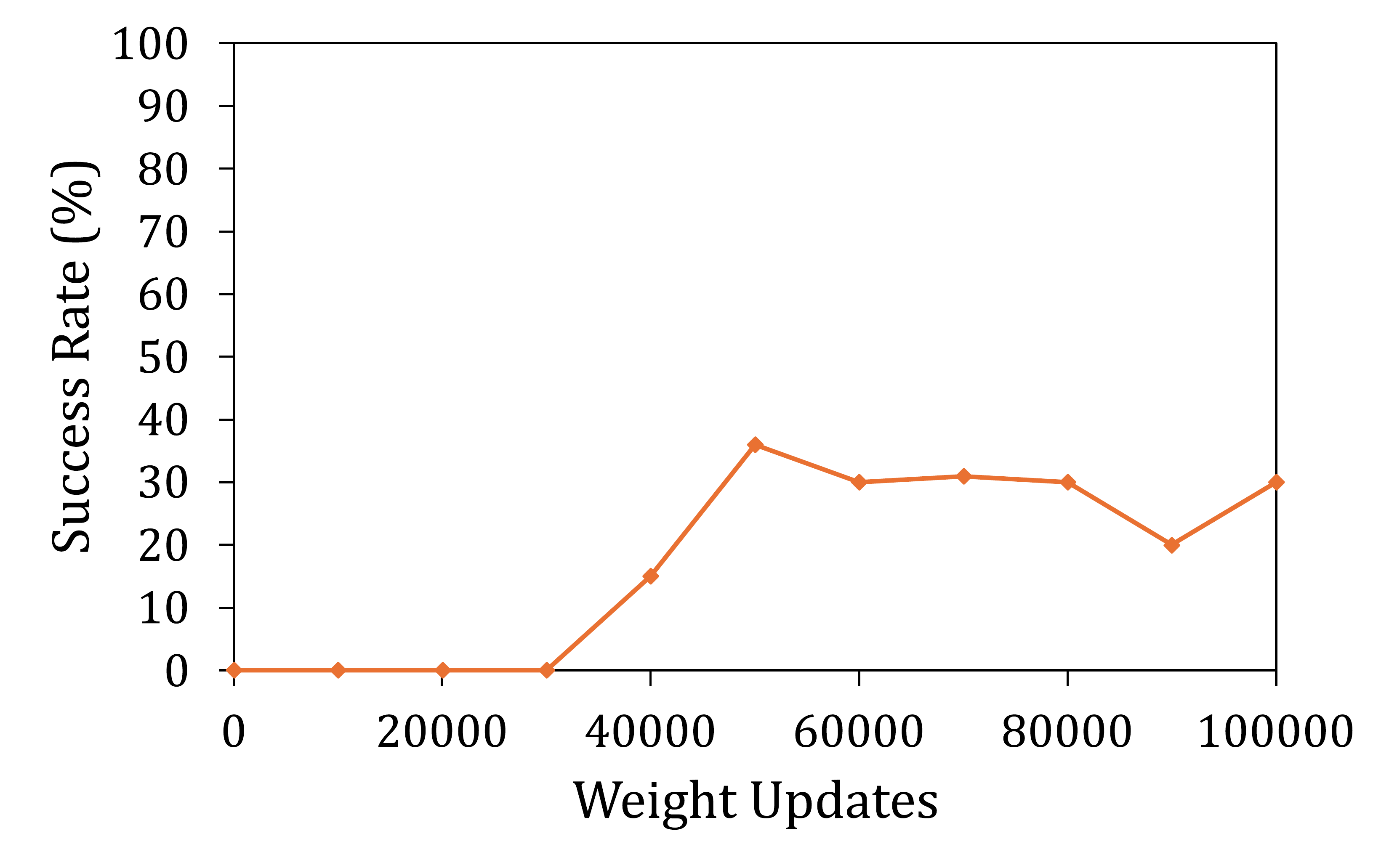}%
      \label{fig:loihi_success_rate}}
    \par\vspace*{0.2cm}
    \rule{0.9\linewidth}{1pt} \\[0.1cm]
    \textbf{\footnotesize Best-Performing Agent}\\[-.15cm]
    \subfigure[]{%
      \includegraphics[width=0.43\linewidth]{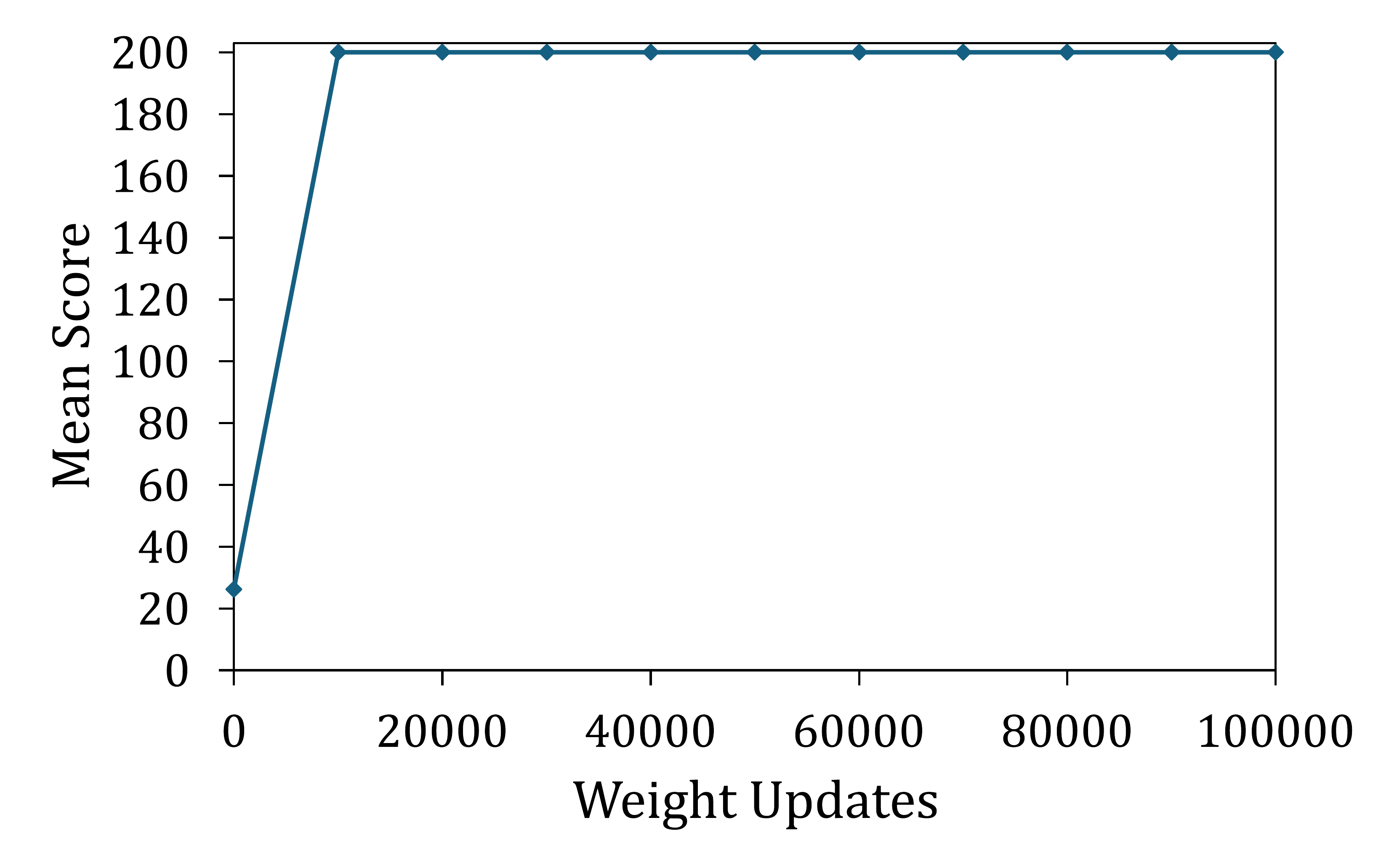}%
      \label{fig:cpu_top_1}}
    \subfigure[]{%
      \includegraphics[width=0.43\linewidth]{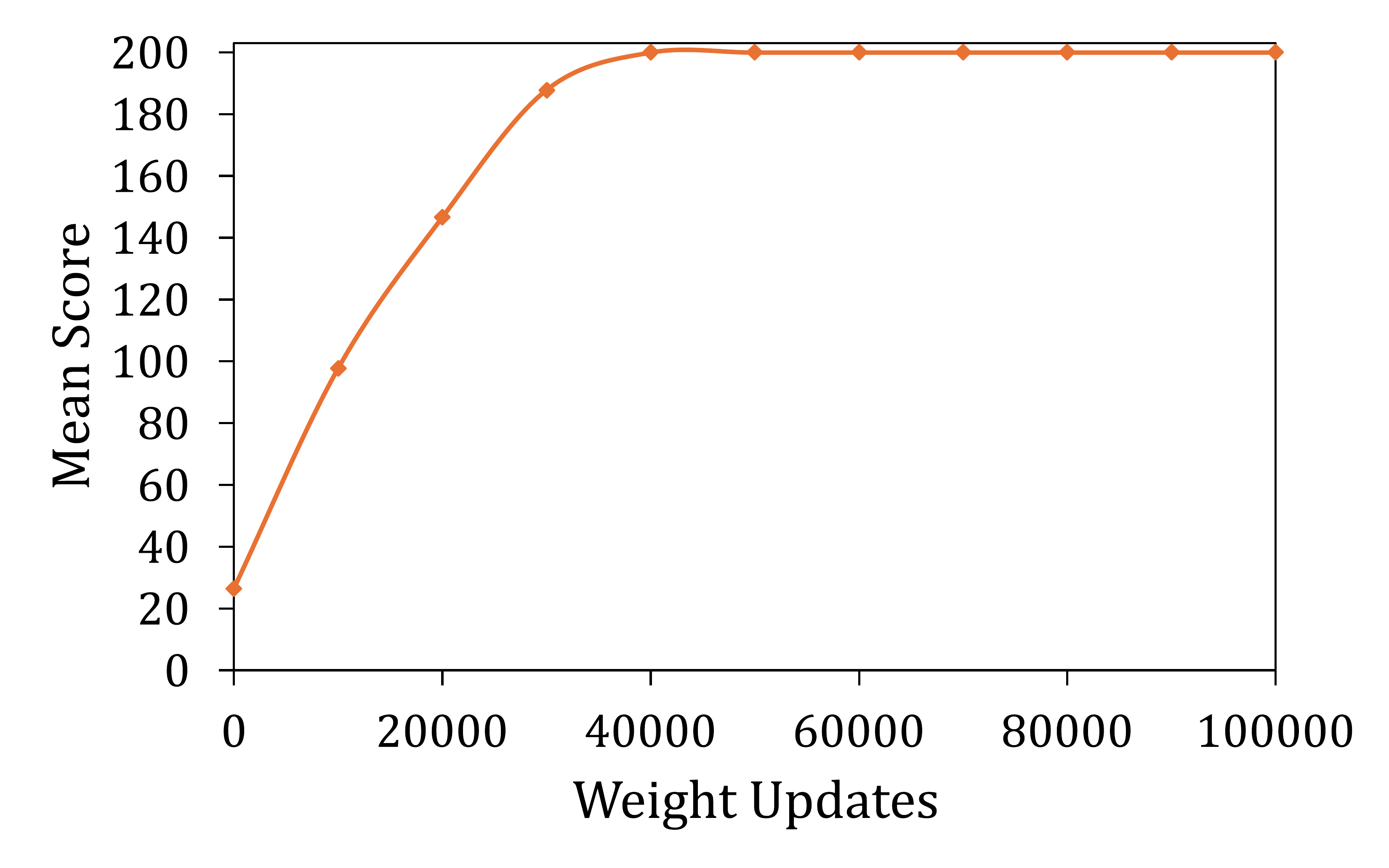}%
      \label{fig:loihi_top_1}} \\[0.1cm]
    \textbf{\footnotesize Top 50\% of Agents}\\[-.15cm]
    \subfigure[]{%
      \includegraphics[width=0.43\linewidth]{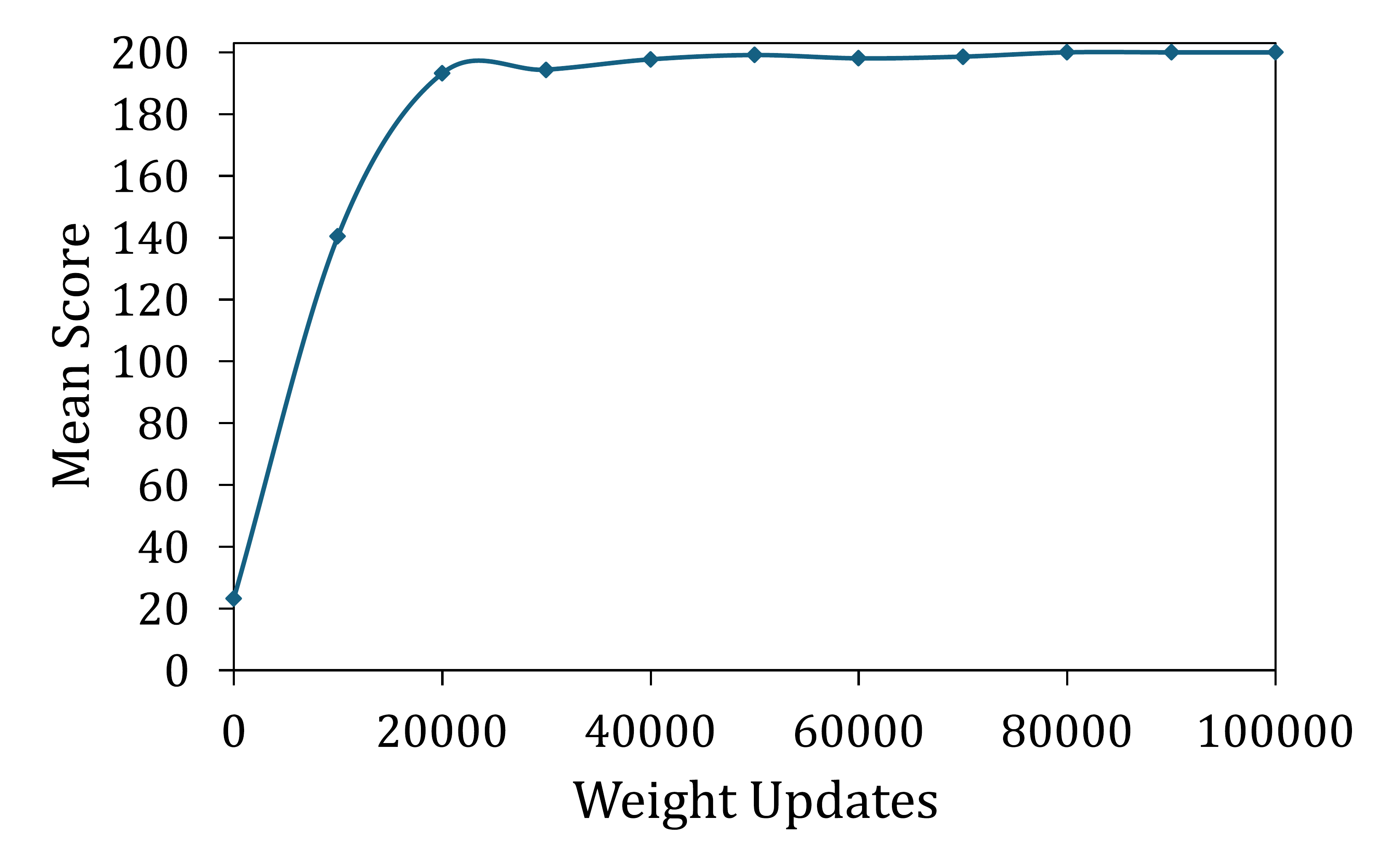}%
      \label{fig:cpu_top_50}}
    \subfigure[]{%
      \includegraphics[width=0.43\linewidth]{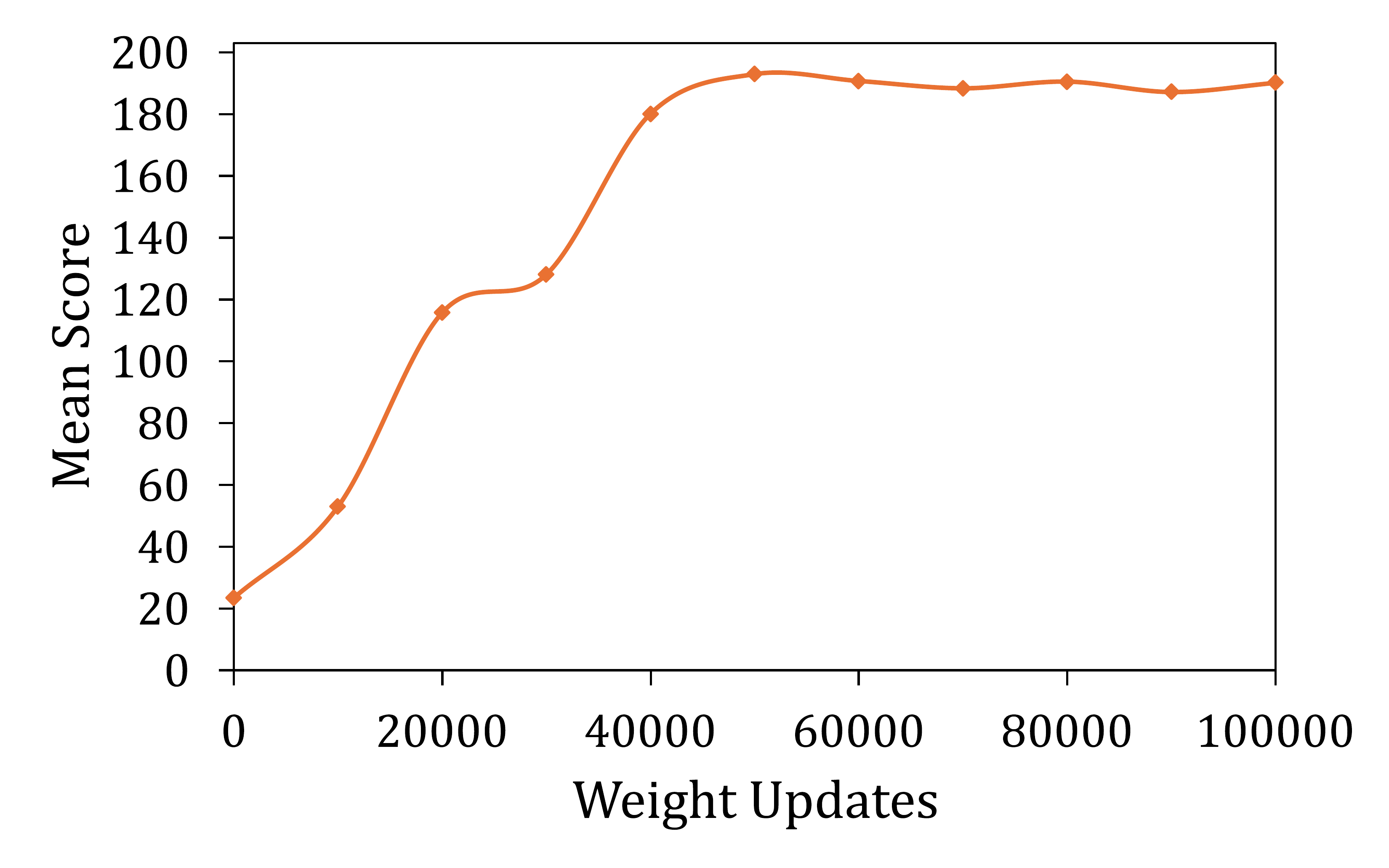}%
      \label{fig:loihi_top_50}} \\[0.01cm]
    \textbf{\footnotesize All Agents}\\[-.15cm]
    \subfigure[]{%
      \includegraphics[width=0.43\linewidth]{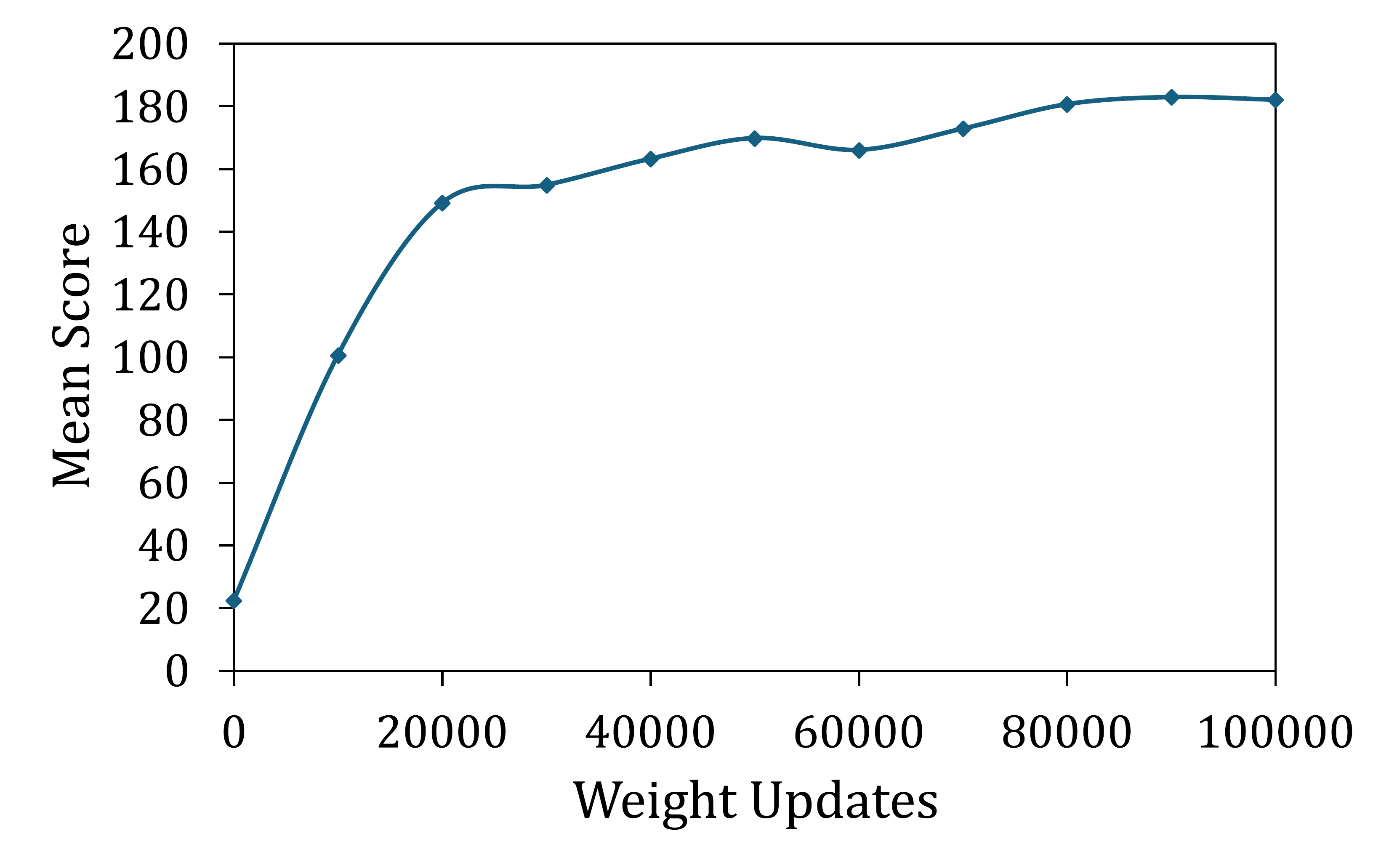}%
      \label{fig:cpu_all_agents}}
    \subfigure[]{%
      \includegraphics[width=0.43\linewidth]{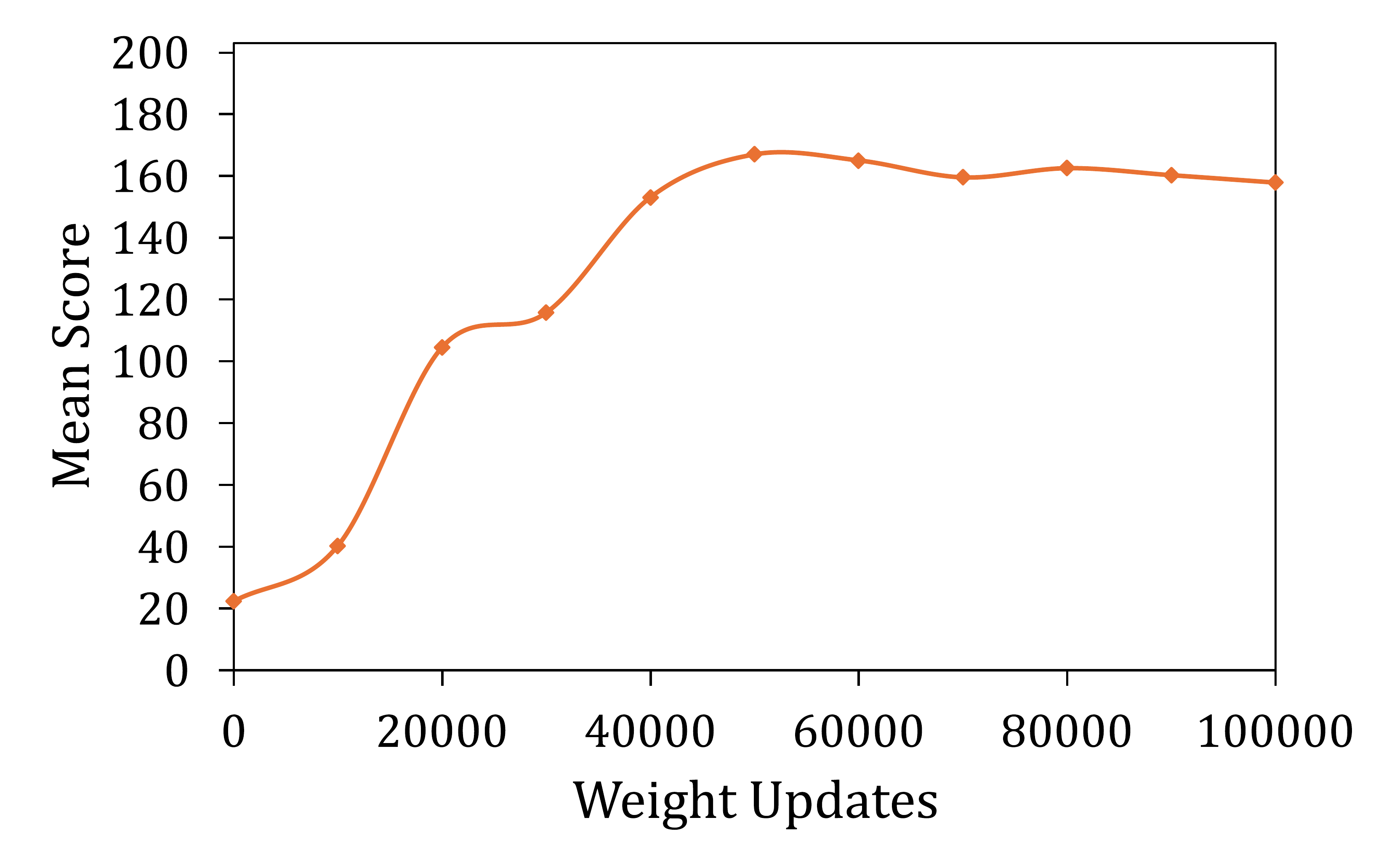}%
      \label{fig:loihi_all_agents}}
    \caption{\textbf{Agent performance on CartPole-v0 after Q-learning on CPU and Loihi~2.} \textbf{a}:~percentage of CPU-trained agents scoring 195 or higher, \textbf{b}:~percentage of Loihi~2-trained agents scoring 195 or higher, \textbf{c}:~CPU-trained top-performing agent, \textbf{d}:~Loihi~2-trained top-performing agent, \textbf{e}:~CPU-trained top 50\% of agents, \textbf{f}:~Loihi~2-trained top 50\% of agents, \textbf{g}:~all CPU-trained agents, \textbf{h}:~all Loihi~2-trained agents.}
      \label{fig:agent_performance}
  \end{center}
\end{figure}

\section{Results}

\subsection{Agent Performance}

To evaluate our neuromorphic Q-learning algorithm against a standard implementation, we trained 100 agents at each of 10 different weight update counts (10\,000 to 100\,000 in increments of 10\,000) on both Loihi~2 and an Intel Core i7-10870H CPU, yielding 1\,000 agents per platform. Both implementations used hyperparameters described in Table~\ref{tab:hyperparameters}. The resulting Q-table matrices were used for inference in the OpenAI Gym CartPole-v0 environment. In CartPole-v0, the agent's score corresponds to the number of cumulative time steps during which the pole remains balanced with the cart in bounds. An agent is considered to have solved the CartPole-v0 task if it achieves a mean score of at least 195 over 100 trials, where each trial is capped at a maximum score of 200. Agent performance across both platforms is shown in Fig.~\ref{fig:agent_performance}---compared against pure random action selection, displayed as agents trained for 0 weight updates.

\subsubsection{CPU Training}

Visualizations of CPU-trained agents are shown in Figs.~\ref{fig:vid_cartpole_python_cpu_trained_0} and \ref{fig:vid_cartpole_python_cpu_trained_1}. The maximum success rate, defined as the proportion of agents able to solve CartPole-v0, achieved during CPU training was 62\%, as shown in Fig.~\ref{fig:cpu_success_rate}. This peak was reached at 90\,000 weight updates and slightly decreased to 61\% at 100\,000 updates. The highest-performing CPU-trained agent attained a perfect mean score of 200 on CartPole-v0 after just 10\,000 weight updates, as depicted in Fig.~\ref{fig:cpu_top_1}. The top 50\% of CPU-trained agents achieved a mean score of 197.74 after 40\,000 weight updates, as shown in Fig.~\ref{fig:cpu_top_50}. Their maximum average score reached a perfect 200 at 80\,000 and 90\,000 weight updates. Similarly, the average score across all CPU-trained agents peaked at 90\,000 updates, reaching a value of 182.94, as shown in Fig.~\ref{fig:cpu_all_agents}.

\subsubsection{Loihi~2 Training}

Visualizations of Loihi~2-trained agents are shown in Figs.~\ref{fig:vid_cartpole_python_on_chip_trained_0} through \ref{fig:vid_cartpole_loihi_on_chip_trained_1}, with visualizations during different stages of training shown in Figs.~\ref{fig:vid_cartpole_loihi_inf_0ts} through \ref{fig:vid_cartpole_loihi_inf_360kts}. The on-chip CartPole environment during training is shown in Figs.~\ref{fig:vid_cartpole_loihi_on_chip_0ts} through \ref{fig:vid_cartpole_loihi_on_chip_360kts}, with various levels of random action selection demonstrated in Figs.~\ref{fig:vid_cartpole_loihi_rand_percent_30} and \ref{fig:vid_cartpole_loihi_rand_percent_10}.

The maximum success rate achieved during training on Loihi~2 was 36\%, as shown in Fig.~\ref{fig:loihi_success_rate}. This peak occurred at 50\,000 weight updates, after which the success rate plateaued near 30\% from 60\,000 updates onward. This success rate, while lower than the CPU's 62\%, represents the first demonstration of successful end-to-end RL training on neuromorphic hardware and achieves the same absolute number of successful agents in half the wall-clock time (see Section~\ref{sec:profiling}). The higher CPU success rate likely stems from several factors: (1) 64-bit floating-point precision vs. 24-bit fixed-point on Loihi~2, (2) much higher trigonometric function accuracy (14--15 decimal places vs. 2--3), and (3) higher-quality random number generation.

Fig.~\ref{fig:loihi_top_1} illustrates the performance of the best Loihi~2-trained agent, which successfully solved the CartPole-v0 environment after 40\,000 weight updates, attaining a perfect mean score of 200 over 100 trials. As shown in Fig.~\ref{fig:loihi_top_50}, peak performance among the top 50\% of agents was observed at 50\,000 updates, at which point their mean score was 193.01. Fig.~\ref{fig:loihi_all_agents} shows the mean scores across all Loihi~2-trained agents. As with the other metrics, the peak average score was obtained at 50\,000 weight updates and reached 167.03. Notably, CPU training peaks later (90\,000 updates) than Loihi~2 (50\,000 updates), suggesting the neuromorphic system may converge faster initially but reach a lower final performance ceiling due to precision limitations. Overall, the neuromorphic Q-learning algorithm achieved strong results in solving the CartPole-v0 task.


\begin{table}[htbp]
\caption{Breakdown of performance profiling for Q-learning training and inference on Loihi~2 and the CPU.}
\centering
\begin{tabular}{|l|l|r|r|r|r|}
\hline
\multirowcell{2}[0pt][c]{\textbf{Task}}
& \multirowcell{2}[0pt][c]{\textbf{Hardware}}&\multicolumn{3}{c|}{\textbf{Power (W)}} & \makecell[c]{\textbf{Latency}}\\
\cline{3-5} 
&& \textbf{\textit{Static}} & \textbf{\textit{Dynamic}} & \textbf{\textit{Total}} & \makecell[c]{\textbf{(\textit{\textmu}s)}}\\
\hline
\multirowcell{3}[0pt][l]{Loihi~2\\training}
& x86 cores & 1.56  & 0.00 & 1.56 &\\
& \makecell[l]{neurocores (34)} & 0.22 & 0.02 & 0.25 &\\
& whole board & 1.78 & 0.02 & 1.80 & 58.42\\
\hline
\makecell[l]{CPU\\training} & \makecell[l]{Intel Core\\i7-10870H} & 31.23 & 12.00 & 43.22 & 162.54\\
\hline
\multirowcell{3}[0pt][l]{Loihi~2\\inference}
& x86 cores & 1.56 & 0.00 & 1.56 &\\
& \makecell[l]{neurocores (12)} & 0.22 & 0.03 & 0.25 &\\
& whole board & 1.78 & 0.03 & 1.81 & 12.86\\
\hline
\makecell[l]{CPU\\inference} & \makecell[l]{Intel Core\\i7-10870H} & 31.64 & 11.63 & 43.27 & 128.55\\
\hline
\end{tabular}\\[0.5cm]
\begin{tabular}{|l|l|>{\raggedleft\arraybackslash}m{1.25cm}|>{\raggedleft\arraybackslash}m{1.25cm}|>{\raggedleft\arraybackslash}m{1.25cm}|}
\hline
\multirowcell{2}[0pt][c]{\textbf{Task}}
& \multirowcell{2}[0pt][c]{\textbf{Hardware}} & \multicolumn{3}{c|}{\textbf{Energy per action (\textit{\textmu}J)}}\\
\cline{3-5} 
&& \makecell[c]{\textbf{\textit{Static}}} & \makecell[c]{\textbf{\textit{Dynamic}}} & \makecell[c]{\textbf{\textit{Total}}}\\
\hline
\multirowcell{3}[0pt][l]{Loihi~2\\training}
& x86 cores & 90.51 & 0.00 & 90.51\\
& \makecell[l]{neurocores (34)} & 12.93 & 1.32 & 14.25\\
& whole board & 103.44 & 1.32 & 104.75\\
\hline
\makecell[l]{CPU\\training} & \makecell[l]{Intel Core\\i7-10870H} & 5\,077.39 & 1\,947.95 & 7\,024.44\\
\hline
\multirowcell{3}[0pt][l]{Loihi~2\\inference}
& x86 cores & 20.01 & 0.00 & 20.01\\
& \makecell[l]{neurocores (12)} & 2.86 & 0.36 & 3.22\\
& whole board & 22.87 & 0.36 & 23.23\\
\hline
\makecell[l]{CPU\\inference} & \makecell[l]{Intel Core\\i7-10870H} & 4\,067.77 & 1\,493.79 & 5\,561.56\\
\hline
\end{tabular}
\label{tab:profiling}
\end{table}

\subsection{Performance Profiling}
\label{sec:profiling}



Table~\ref{tab:profiling} shows performance profiling for Q-learning on Loihi~2 and the CPU. Measurements for the CPU were conducted using HWiNFO64 version 8.26-5730 on a Dell Alienware m15 R4 laptop running Microsoft Windows 10 Enterprise version 10.0.19045 Build 19045. The system was equipped with an Intel Core i7-10870H CPU operating at 2.20 GHz (2208 MHz), with 8 physical cores and 16 logical processors. Measurements for Loihi~2 were performed using Lava on Loihi version v0.6.0, running on Oheogulch board \texttt{ncl-ext-og-05}, generation N3B3, and recorded from on-chip energy and execution time probes. The Loihi~2 board contains both neurocores and x86 cores, so measurements are reported separately for each of these hardware components. Energy per action (bottom half of Table~\ref{tab:profiling}) was calculated by multiplying power by latency.

Table~\ref{tab:profiling} reveals that Loihi~2's energy advantage stems from both lower power draw (\(600\times\) during training) and faster execution (\(2.8\times\) per weight update), compounding to yield over \(1\,000\times\) energy efficiency for equivalent training success. Given that CPU training takes 2.8 times longer per weight update, Loihi~2 remains faster in terms of wall-clock time despite its lower success rate; requiring only 51\% as much time to train the same number of successful agents.


We do not report GPU performance for CartPole Q-learning because a direct comparison to our neuromorphic implementation would have required a fully on-device environment identical to OpenAI Gym’s CartPole-v0 as well as integer-precision learning updates that match our fixed-point constraints (8-bit weights with 24-bit activations). This would require custom CUDA implementations of both the environment and the Q-updates, which is beyond the scope of this work and is unlikely to utilize the GPU efficiently for this iterative, constrained state space problem. As an example of the inefficiency of using existing libraries, training on an NVIDIA Tesla V100S PCIe 32GB GPU via PyTorch using the OpenAI Gym CartPole-v0 environment with a 67\% success rate resulted in three orders of magnitude higher energy consumption and an order of magnitude longer execution time for the GPU than Loihi~2.



\section{Discussion}

\subsection{Training Algorithm Performance}

This work demonstrates that neuromorphic hardware is not only capable of solving RL tasks like CartPole, but also can do so with energy efficiency over three orders of magnitude greater than traditional CPUs, while matching CPU-trained agent success in half the training time. Prior work has deployed reinforcement-learning policies and value networks on neuromorphic hardware, including Loihi-based robotic control and deep spiking Q-networks \cite{tang2020reinforcement,tang2021population,akl2021porting}. To the authors' knowledge, the distinguishing contribution here is the fully on-chip, closed-loop Loihi~2 implementation in which the Q-learning update and the CartPole simulation execute on the neuromorphic processor without host computation between environment transitions. This model-free implementation therefore extends prior neuromorphic RL deployments from efficient policy execution toward a self-contained learning--environment loop.

\subsection{Limitations}

The neuromorphic implementation achieves 58\% of the CPU's success rate (36\% vs. 62\%) likely due to Loihi~2's 24-bit fixed-point arithmetic compared to the CPU's 64-bit floating-point arithmetic and reduced RNG performance in random action selection. Future work should investigate whether this gap can be narrowed through improved learning rate schedules adapted to fixed-point arithmetic, alternative \mbox{Q-table} representations, or improved RNG. While an LCG was necessary for use within the CartPole neuron, higher-quality generators such as Wichmann-Hill \cite{wichmann_hill} could be used within the random action selection circuitry to improve state space exploration \cite{rng_comparison}.

CartPole serves only as an initial benchmark for neuromorphic RL. More complex control tasks, such as unmanned aerial system (UAS) control~\cite{diego1,diego2}, would provide a more realistic assessment of real-world applicability and appear amenable to on-chip implementation, as suggested by our single-neuron representation of the CartPole environment. Demonstrating on-chip learning for such tasks would indicate that agents can be continuously adapted \emph{in situ} without disrupting task execution, since our implementation already sustains over 17~kHz weight-update rates, comfortably above the 100--1\,000~Hz sensing and actuation frequencies typical for quadcopter control~\cite{quadcopter}.


\subsection{Broad Implications}


For the class of control tasks with discrete state--action spaces and real-time requirements---such as robotics, autonomous systems, and edge computing---this work demonstrates that neuromorphic RL offers a viable path toward \(1\,000\times\) energy efficiency improvements. Real impact on AI sustainability will, however, require showing that such gains extend to more complex benchmarks and ultimately to deployed applications.

Taken together with neuromorphic backpropagation~\cite{spikingbackprop}, these results show that the synfire-gated synfire chain~\cite{synfire} coding framework can support a broad class of AI algorithms previously realized only on von Neumann architectures. Algorithms originally designed for globally-clocked processors can now be mapped onto asynchronous neuromorphic chips, suggesting that larger, more capable neuromorphic AI models are feasible and that this framework offers a plausible path toward energy-efficient, commercially relevant AI systems.

\section*{Acknowledgments}

We thank Alpha Renner for valuable discussions and acknowledge Intel Labs and the INRC community for providing access to the Loihi research chip and for discussions and support. This work has been approved for unlimited release by Los Alamos National Laboratory with number LA-UR-25-31518.


\balance
\bibliography{references}

\end{document}